\documentclass{article}

\usepackage{microtype}
\usepackage{graphicx}
\usepackage{subfigure}
\usepackage{booktabs} % for professional tables
\usepackage{wrapfig}
\usepackage{tcolorbox}
\usepackage{hyperref}

\usepackage{multirow}

\usepackage[accepted]{icml2025}

\usepackage{amsmath}
\usepackage{amssymb}
\usepackage{mathtools}
\usepackage{amsthm}

\usepackage[capitalize,noabbrev]{cleveref}

\theoremstyle{plain}
\newtheorem{theorem}{Theorem}[section]

\theoremstyle{definition}
\newtheorem{definition}[theorem]{Definition}

\theoremstyle{remark}
\newtheorem{remark}[theorem]{Remark}
\newtheorem{example}[theorem]{Example}

\usepackage[textsize=tiny]{todonotes}

\icmltitlerunning{Structured Captions Improve Prompt Adherence in Text-to-Image Models (Re-LAION-Caption 19M)}

\begin{document}

\onecolumn

\icmltitle{Structured Captions Improve Prompt Adherence in Text-to-Image Models (Re-LAION-Caption 19M)}

% It is OKAY to include author information, even for blind
% submissions: the style file will automatically remove it for you
% unless you've provided the [accepted] option to the icml2025
% package.

% List of affiliations: The first argument should be a (short)
% identifier you will use later to specify author affiliations
% Academic affiliations should list Department, University, City, Region, Country
% Industry affiliations should list Company, City, Region, Country

% You can specify symbols, otherwise they are numbered in order.
% Ideally, you should not use this facility. Affiliations will be numbered
% in order of appearance and this is the preferred way.
\icmlsetsymbol{equal}{*}

\begin{icmlauthorlist}
\icmlauthor{Nicholas Merchant}{equal,supermodel}
\icmlauthor{Haitz Sáez de Ocáriz Borde}{equal,supermodel}
\icmlauthor{Andrei Cristian Popescu}{equal,supermodel}
% \icmlauthor{Pantelis Papageorgiou}{supermodel}
\icmlauthor{Carlos Garcia Jurado Suarez}{supermodel}
% \icmlauthor{Anonymous Author}{Anonymous}
%\icmlauthor{}{sch}
%\icmlauthor{}{sch}
\end{icmlauthorlist}

\icmlaffiliation{supermodel}{Supermodel}
\icmlcorrespondingauthor{Haitz Sáez de Ocáriz Borde}{ocariz@supermodel.ai}

% You may provide any keywords that you
% find helpful for describing your paper; these are used to populate
% the "keywords" metadata in the PDF but will not be shown in the document
\icmlkeywords{}

\vskip 0.3in

% this must go after the closing bracket ] following \twocolumn[ ...

% This command actually creates the footnote in the first column
% listing the affiliations and the copyright notice.
% The command takes one argument, which is text to display at the start of the footnote.
% The \icmlEqualContribution command is standard text for equal contribution.
% Remove it (just {}) if you do not need this facility.

%\printAffiliationsAndNotice{}  % leave blank if no need to mention equal contribution
\printAffiliationsAndNotice{\icmlEqualContribution} % otherwise use the standard text.

\begin{figure*}[ht!]
    \centering
\includegraphics[width=0.95\textwidth,trim= 0 30 0 0,clip]{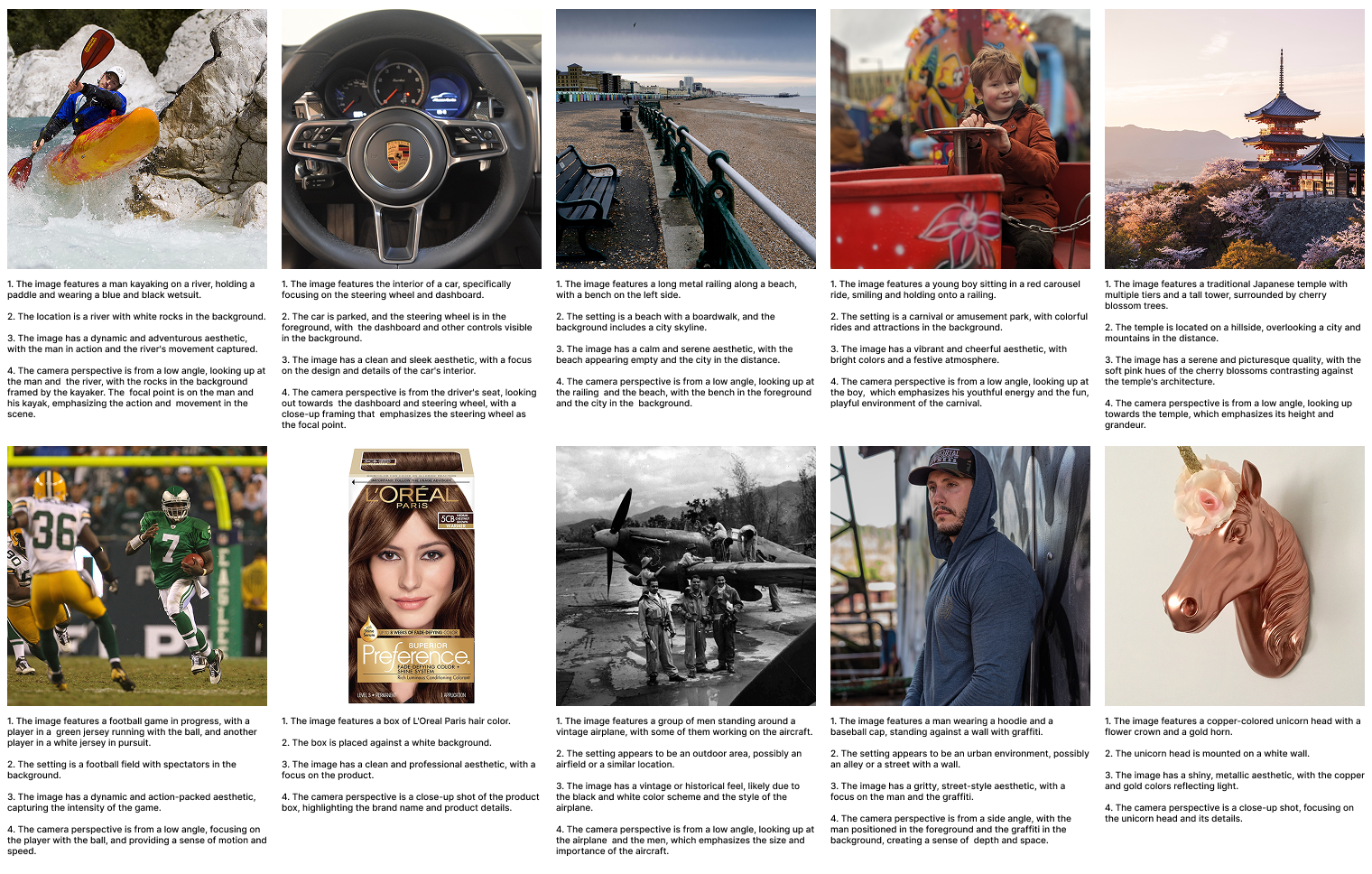}
    \caption{Images from Re-LAION-Caption 19M with structured captions: 1. subject, 2. setting, 3. aesthetics, and 4. camera details.}
    \label{fig:banner_image}
\end{figure*}

\begin{abstract}
We argue that generative text-to-image models often struggle with prompt adherence due to the noisy and unstructured nature of large-scale datasets like LAION-5B. This forces users to rely heavily on prompt engineering to elicit desirable outputs. In this work, we propose that enforcing a consistent caption structure during training can significantly improve model controllability and alignment. We introduce Re-LAION-Caption 19M, a high-quality subset of Re-LAION-5B, comprising 19 million 1024×1024 images with captions generated by a Mistral 7B Instruct-based LLaVA-Next model. Each caption follows a four-part template: subject, setting, aesthetics, and camera details. We fine-tune PixArt-$\Sigma$ and Stable Diffusion 2 using both structured and randomly shuffled captions, and show that structured versions consistently yield higher text-image alignment scores using visual question answering (VQA) models. The dataset is publicly available at~\texttt{https://huggingface.co/datasets/supermodelresearch/Re-LAION-Caption19M}.
\end{abstract}

\clearpage

\twocolumn

\section{Introduction}

A significant portion of the recent performance gains in deep learning models can be attributed to the availability of larger and higher-quality datasets than those available in previous decades. In particular, in the context of text-to-image models, names like DALL-E~\citep{ramesh2021zeroshottexttoimagegeneration} and Stable Diffusion~\citep{rombach2022highresolutionimagesynthesislatent} have greatly benefited from open-source, web-scale datasets such as LAION-5B~\citep{schuhmann2022laion5bopenlargescaledataset}. LAION-5B, as its name suggests, originally consisted of up to 5.85 billion CLIP-filtered~\citep{radford2021learningtransferablevisualmodels} image-text pairs.

At this scale, it becomes notoriously difficult to pre-filter and reliably evaluate the dataset, and indeed several issues were identified a posteriori which forced the takedown of the original dataset and led to the development of Re-LAION-5B, as we discuss in Section~\ref{sec:Re-LAION-5B}. In this work, we focus on the lack of structure in the image descriptions, both in the case of LAION-5B and Re-LAION-5B. To produce the captions, the authors parsed Common Crawl’s WAT metadata to extract every HTML image tag with an alt attribute, treating that alt-text as the raw caption. Each caption was then run through CLD3 language detection~\citep{ooms2022cld3} (English, other, or no language), and any captions under five characters were discarded, yielding the `cleaned' alt-text captions used in the dataset. This procedure offers no control over the prompt structure fed into the generative model, likely making text-to-image learning unnecessarily difficult due to the high variability and irregularity of raw text data found in the wild. In this paper, we argue that the reliability and controllability of generative models are deeply connected to the nature of their training data.

We hypothesize that training (or post-training) text-to-image generative models with consistent prompt structures enhances controllability. Such consistency simplifies the learning process by removing the need for the model to become invariant to different caption formats. Furthermore, a predefined prompt pseudo-template helps users ensure their queries align with the model’s training distribution, reducing the need for extensive prompt engineering at inference time. In Section~\ref{sec:Caption Structure Invariance} we formalize this intuition using group theory, and based on this we propose Re-LAION-Caption 19M a new filtered subset of Re-LAION-5B containing high-resolution 1024×1024
images of high quality and re-captioned using a consistent prompt structure. We generate the new captions with the open-source LLaVA-Next model~\citep{liu2024llavanext} (also known as v1.6) leveraging Mistral 7B Instruct~\citep{jiang2023mistral7b} (see Section~\ref{sec:Re-Captioning with Vision Language Models} for more details).

Other efforts that follow our line of research in the literature include Recap DataComp-1B~\citep{li2024recaptionbillionswebimages}, VeCLIP~\citep{lai2024veclipimprovingcliptraining}, and LaCLIP~\citep{fan2023improvingcliptraininglanguage}, all of which recaption web-scale datasets to improve image-text alignment. However, our work is different in that unlike these methods which use single-sentence captions or unconstrained rewrites, we introduce a structured captioning format that separates the content into four interpretable fields to improve controllability at a more granular level. Also, while Recap and VeCLIP primarily target CLIP-style retrieval or classification benchmarks, our focus is on text-to-image.

\section{Caption Structure Invariance}
\label{sec:Caption Structure Invariance}

In this section, we employ group theory to formalize the intuition behind the benefits of training models with a consistent prompt structure.

Let $\Omega$ denote the set of all English captions. Each caption $c \in \Omega$ is considered as a sequence of tokens, where $c(t)$ represents the token at position $t$. Define the input feature function space:

$$
\mathcal{X}(\Omega) = \{ x : \Omega \to \mathbb{R}^d \},
$$

where each $x \in \mathcal{X}(\Omega)$ maps any caption $c \in \Omega$ to a $d$-dimensional feature representation $x(c) \in \mathbb{R}^d$, typically produced by a text encoder. The dimension $d$ denotes the embedding dimension, distinct from any token-level representation dimension $d'$. This distinction separates the feature vectors obtained at the token-level (such as from the codebook embedding layer) from those obtained after encoding a complete input prompt through multiple Transformer encoder layers.

Define a group $G$ whose elements $g \in G$ represent meaning-preserving transformations acting on captions.

\begin{definition}[Meaning-Preserving Group Action]
A group action $\rho : G \times \Omega \to \Omega$ mapping each transformation $g \in G$ and caption $c \in \Omega$ to a new caption $\rho(g)c \in \Omega$ with equivalent semantics.
\end{definition}

\begin{example}[Sentence Shuffling]
A specific instance of such a meaning-preserving transformation is sentence-level shuffling: given a caption composed of multiple sentences, reordering these sentences typically preserves the overall semantic content, even though the structure of the caption changes. In practice one must ensure sentences do not carry forward-referencing dependencies for this to hold. This is particularly true when the sentences are self-contained, as in our dataset, see Section~\ref{sec:Re-Captioning with Vision Language Models}. 
\end{example}

\begin{remark}[Retokenization]
Note that technically $\rho(g)c$ need not merely permute the tokens in $c$. Some $g \in G$ may indeed represent permutations (thus $(\rho(g_p)c)(t) = c(p^{-1}(t))$), but generally, transformations in $G$ can yield entirely new sequences of tokens, since changing the order of words or sentences may cause the tokenizer to produce a different tokenization rather than simply rearranging the original tokens ($(\rho(g_p)c)(t) \neq c(p^{-1}(t))$). Remember that most text tokenizers will re-encode the shuffled text from scratch.
\end{remark}

The action of $G$ on $\Omega$ induces an action on $\mathcal{X}(\Omega)$. For $x \in \mathcal{X}(\Omega)$ and $g \in G$, define:

$$
(g \cdot x)(c') = x(\rho(g^{-1})c'), \quad c' \in \Omega.
$$

In other words, the group action on captions induces a natural action on the functions defined over captions. Rather than acting directly on the output vectors, the group transforms the input caption before evaluating the function.

\begin{definition}[Caption Structure Invariance]
A model $f : \mathcal{X}(\Omega) \to \mathcal{Y}$ is \emph{caption structure-invariant} if:

$$
f(g \cdot x) = f(x), \quad \forall g \in G, \forall x \in \mathcal{X}(\Omega).
$$

\end{definition}

Instead of enforcing invariance universally, we propose restricting attention to a structured subset $\Omega' \subset \Omega$, generated using a fixed prompt template. Let $u \in \mathcal{U}$ denote semantic content variables drawn from an allowed set $\mathcal{U}$. A caption $c_{\text{template}}(u) \in \Omega'$ is then constructed via a fixed template, for example:

$$
c_{\text{template}}(u) = ||(s_1, \phi_1(u), s_2, \dots, s_m, \phi_m(u)), \quad u \in \mathcal{U},
$$

where fixed token sequences $s_j$ (in Section~\ref{sec:Re-Captioning with Vision Language Models} this would be ``1.'',``2.'',``3.'',``4.'') interleave content-derived sequences $\phi_j(u)$, and $||$ denotes concatenation. The set $\Omega'$ defines the \emph{canonical caption structure domain}:

\begin{definition}[Canonical Caption Structure Domain $\Omega'$]
The canonical caption structure domain $\Omega'$ comprises captions generated by a fixed template, satisfying invariance under $G$, i.e.,

$$
\forall g \in G, \forall c \in \Omega' : \rho(g)c = c.
$$

\end{definition}

Thus, we restrict to the subspace $\mathcal{X}(\Omega') \subset \mathcal{X}(\Omega)$, where each $x \in \mathcal{X}(\Omega')$ primarily operates within $\Omega'$. Consequently, the model $f$ need only learn:

$$
f|_{\mathcal{X}(\Omega')} : \mathcal{X}(\Omega') \to \mathcal{Y},
$$

a considerably simpler task compared to achieving invariance across all $\Omega$. Since each caption in $\Omega'$ is canonical, invariance trivially holds. For any $x \in \mathcal{X}(\Omega')$ and $c' \in \Omega'$, we have:

$$
(g \cdot x)(c') = x(\rho(g^{-1})c') = x(c'), \quad \text{and thus} \quad g \cdot x = x.
$$

This dramatically simplifies the invariance requirement on $f$: the problem is shifted from complex learning to careful input design. Note that because we have defined the canonical caption domain $\Omega'$ explicitly as invariant under all transformations in $G$, the caption-structure invariance property trivially follows. Here, our intention is to mathematically and clearly illustrate the simplicity gained from adopting structured caption templates compared to dealing with general, unstructured captions.

The benefits derived from this approach are immediate:
\begin{itemize}
\item \textbf{Coverage}: Maintains representation of all essential concepts (e.g., objects and styles).
\item \textbf{Simplicity}: Eliminates combinatorial syntax variability by adopting a structured slot sequence.
\item \textbf{Efficiency}: Redirects model capacity toward semantic learning rather than invariance to structural permutations.
\end{itemize}

While the group-theoretic framework does not introduce new algorithmic techniques, it provides a formal lens for understanding how structure in caption templates can offload the burden of learning invariance from the model to the data. This helps articulate a design principle for dataset construction.

\section{LAION-5B and Re-LAION-5B}
\label{sec:Re-LAION-5B}

In this section, we describe the web-scale dataset used to construct our subset.

\paragraph{LAION-5B} LAION-5B was constructed by crawling publicly available web pages from Common Crawl to extract image URLs paired with their author-provided alt text. These images were then downloaded, and any that were too small or malformed were discarded. The remaining image-text pairs were encoded using CLIP models to compute cosine similarity. Low-confidence matches (about $90\%$ of the data) were filtered out, retaining approximately 5.85 billion high-quality examples. This corpus was subsequently split into English ($2.32$ billion), multilingual ($2.26$ billion), and no-language ($1.27$ billion) subsets. Additional automatic annotations for adult content, watermarks, and toxicity were added for downstream filtering~\citep{schuhmann2022laion5bopenlargescaledataset}. Despite this initial effort, LAION-5B has been reported to have multiple issues, such as an estimated 700 million duplicate images~\cite{webster2023deduplicationlaion2b} as well as copyright conflicts.

\paragraph{Re-LAION-5B} Following a safety revision prompted by the discovery of links to suspected CSAM within the original LAION-5B dataset, LAION e.V. recently (30 Aug, 2024) released Re-LAION-5B, a cleaned version. This updated dataset, created in partnership with the Internet Watch Foundation (IWF), the Canadian Center for Child Protection (C3P), and Stanford Internet Observatory, involved removing 2236 identified links by matching against over 16 million image and URL hashes of known illegal content, without directly accessing the potentially harmful material. Re-LAION-5B, containing approximately 5.5 billion text-image pairs, aims to provide a safer and more reliable open dataset for reproducible language-vision learning research, while emphasizing the limitations of automated filtering and recommending collaboration with expert organizations for improved safety in web-scale dataset creation.

\section{Filtering Scores} 
\label{sec:Filtering Scores}

Next we describe our dataset filtering procedure: concretely, we work with the Re-LAION-5B research safe English 2 billion subset\footnote{\texttt{relaion2B-en-research-safe} accessible at https://huggingface.co/datasets/laion/relaion2B-en-research-safe}. 

\paragraph{Initial Size and Aspect Ratio Filtering} First, we discard images whose width or height is less than 1024 pixels. This ensures a minimum resolution for input images. Second, we filter out images with extreme aspect ratios. The code calculates the minimum of the width/height and height/width ratios and keeps images only if this ratio is greater than or equal to 0.6666. This means we accept images where the shorter side is at least $66.66\%$ of the length of the longer side, effectively selecting images that are reasonably close to square. Finally, we center crop these images at 1024×1024. This pre-filtering effectively reduces the initial dataset to 39,149,128 images.

\paragraph{Score-based Filterings} After this initial step we proceed to filter and classify the images according to different quality metrics:

\begin{itemize}
    \item \textit{Aesthetic Score:} rates the quality of the image according to a pre-trained open-source model that simulates human preference (this model is a combination of an OpenAI CLIP ViT L14 model~\citep{radford2021learningtransferablevisualmodels} and MultiLayer Perceptron~(MLP) aesthetic score predictor\footnote{Accessible at https://github.com/christophschuhmann/improved-aesthetic-predictor which is referred to as \texttt{LAION-Aesthetics\_Predictor V2}. This is the same model used to prepare the LAION subset LAION Aesthetic V2 from https://laion.ai/blog/laion-aesthetics/}). We keep images with an aesthetic score greater than 4.73.
    \item \textit{Luminance:} following OpenHumanVid ~\citep{li2025openhumanvidlargescalehighqualitydataset}, we calculate the mean luminance of an image based on the following formula:
    \begin{equation}
    L = \frac{1}{N} \sum_{i=1}^{N} \left( r \cdot R_i + g \cdot G_i  + b \cdot B_i \right)
\end{equation}
    with the scalars $r=0.2126, g=0.7152, b=0.0722$, and where $R_i$, $G_i$, and $B_i$ correspond to the red, green, and blue channels of the $i$-th pixel, respectively, and $N$ is the total number of pixels for a given image. We keep samples with a luminance score $L\in [12.75, 204.00]$ to filter out images that are too bright or too dark, as well as to minimize the presence of only product store images.
    \item \textit{Text detection by custom OCR score:} images are analyzed for the presence of text using a custom Optical Character Recognition (OCR) score, which takes values between 0 and 1. The OCR score is computed using a Differentiable Binarization (DB) model \citep{liao2019realtimescenetextdetection} with a ResNet-18 backbone. The DB model predicts polygons enclosing detected text regions and associated confidence scores for each polygon. Next, for each detected polygon, its area in pixel units is calculated. The OCR score for each image is computed as follows:
    \begin{equation}
        \text{OCR Score} = \frac{\sum_{p \in \mathcal{P}} (\text{A}(p) \times \text{C}(p))}{S^2},
    \end{equation}
    where $\mathcal{P}$ is the set of all detected text polygons in the image with a confidence score greater than or equal to 0.7\footnote{The threshold is based on recommendations in https://github.com/MhLiao/DB}; $\text{A}(p)$ is the area (in pixels) of the individual polygon $p$; $\text{C}(p)\in[0.7,1]$ is the confidence score associated with the detected polygon $p$; $S$ is the side length of the square input image (736 pixels in this implementation: input images are resized to 736x736 resolution), so $S^2$ represents the total area of the input image. This custom OCR score represents the weighted proportion of the image area covered by confidently detected text regions. A higher score indicates a greater presence of text with higher confidence. This metric is used in our filtering process to select images based on the amount and reliability of detected text. We keep images with both very low (most of the dataset) and very high OCR scores. This is because intermediate OCR scores around 0.1 to 0.6 primarily include scribbles or text too small, which we find to be undetectable or confusing for LLaVa models as well as generally hard to reconstruct with autoencoders. Hence we would generally like to aim for images that either do not contain text or that feature sufficiently large and clear text beneficial for downstream training.
\end{itemize}

All filtering steps combined reduce the image count from 39,149,128 to 19,055,277. We provide a logbook of our filtering procedure in Appendix~\ref{sec:Logbook}. We also provide visualizations before and after filtering the dataset in Appendix~\ref{sec:Visualizing}. Finally, note that we deem blur filtering unnecessary in our case, as we have already restricted the subset to high-resolution images. While some images exhibit blur, this is often a feature rather than a bug; for instance, blurred backgrounds can be a stylistic choice to emphasize the main subject.

\section{Structured Re-Captioning with LLaVa-Next}
\label{sec:Re-Captioning with Vision Language Models}

For re-captioning, we use LLaVa-Next~\citep{liu2024llavanext} with Mistral 7B Instruct. We choose this model to balance between caption quality and inference speed as compared to other models in the LLaVa-Next family. We use the following system prompt:

\begin{tcolorbox}[colback=gray!10,colframe=black,arc=4mm]
\texttt{Describe the image using the following structure (4 sentences in total, using bullet points): 1. Subjects or objects in the image in one sentence, including actions if applicable. 2. Location and setting in one sentence. 3. Image aesthetics in one sentence. 4. Camera perspective, including angle, framing, and focal point details in one sentence.}
\end{tcolorbox}

\begin{figure}[htbp!]
    \centering
\includegraphics[width=0.4\textwidth]{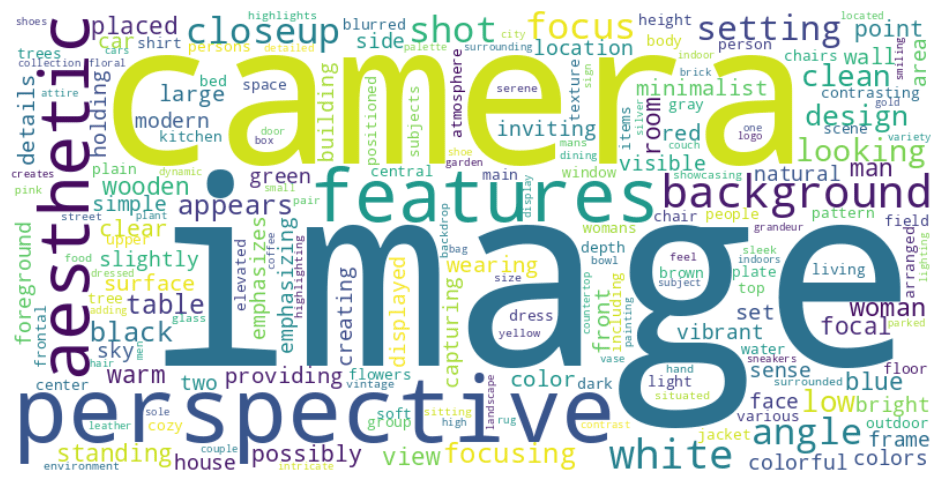}    \caption{Most common words produced by LLaVa-Next during re-captioning, excluding stopwords. Note that ``camera'', ``image'', ``aesthetic'', and ``perspective'' appear often, as expected given the system prompt.}
    \label{fig:worldcloud}
\end{figure}

Using bullet points helps the model clearly separate sentences corresponding to different tasks. For the first task, we observed the model occasionally uses two sentences instead of one, see for instance the first bullet point in the unshuffled caption of Figure~\ref{fig:shuf_vs_unshuf}. Furthermore, we found it challenging initially to make the model distinguish between image aesthetics and camera details. Separating these as two distinct requests in the system prompts proved paramount; otherwise, the model would ignore one of them. Notably, using \textit{image aesthetic} rather than \textit{image style} was particularly effective. A word cloud of the most commonly used words by the LLaVa-Next model is displayed in Figure~\ref{fig:worldcloud}.

\paragraph{Defective LLaVa captions} After our image captioning run, we applied a script to determine whether captions adhered to the desired bullet point template format (as illustrated in Figure~\ref{fig:banner_image}) and we removed defective samples, reducing the dataset size from 19,055,277 to 19,038,079 text-image pairs. Failure cases are discussed in Appendix~\ref{sec:LLaVa captioning failure cases}. Another problem that we observe regarding the captions produced by LLaVa-Next is that it likes to repeat itself when describing aesthetics and camera details: it often mentions again redundant information present in bullet points 1. or 2. We found this hard to remove programmatically.

\section{Text-to-Image Fine-tuning Experiments}
\label{sec:Text-to-Image Fine-tuning Experiments}

\begin{figure}[htbp!]
    \centering
    \includegraphics[width=0.2\textwidth]{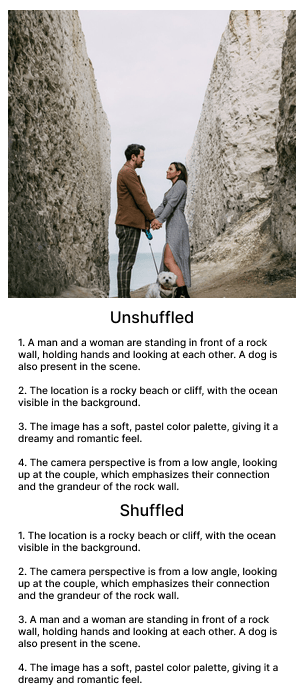}
    \caption{The unshuffled (or structured) caption corresponds to the true caption used in Re-LAION-Caption 19M. The shuffled version is used to test the effect of unstructured captions during fine-tuning.}
    \label{fig:shuf_vs_unshuf}
\end{figure}

To verify our hypothesis that structured captions aid learning and contribute to more robust text-to-image models, we fine-tune PixArt-$\Sigma$~\citep{chen2024pixartsigmaweaktostrongtrainingdiffusion} and Stable Diffusion Version~2~\citep{rombach2022highresolutionimagesynthesislatent} using two variations of the dataset. The first dataset is Re-LAION-Caption 19M, and the second consists of the same set of images but with randomly shuffled caption structures. For instance, as shown in Figure~\ref{fig:shuf_vs_unshuf}, a default caption example follows the subject/object, location/setting, aesthetic, and camera ordering. In contrast, its shuffled version would point to the same image and contain the same net information, but the ordering would be randomized (the permutation is different for each image in the dataset). Note that this is a controlled environment: despite permuting the caption order, both datasets have significantly more structure than what we would typically find in LAION.

\paragraph{Extending the Context Window of CLIP-based Models} PixArt-$\Sigma$ utilizes a 4.3B Flan-T5-XXL~\citep{chung2022scalinginstructionfinetunedlanguagemodels} text encoder with a maximum context length of 300 tokens, which is sufficient for nearly all of our dataset, with the exception of 33 captions. In contrast, Stable Diffusion employs CLIP-based encoders with a maximum context length of only 77 tokens. To enable fine-tuning on our captions, we extend the text encoders by concatenating T5 text encoder embeddings with their original CLIP embeddings along the sequence dimension. We found removing the original CLIP embeddings to be excessively aggressive.

\paragraph{Attending to the Tokenizer} 
Our generative models are based on T5~\citep{raffel2023exploringlimitstransferlearning} and Flan-T5~\citep{chung2022scalinginstructionfinetunedlanguagemodels}. To accommodate their tokenizer, we remove newline characters from strings produced by LLaVa-Next, preventing potential confusion. Additionally, we replace the numbering format ``1.'', ``2.'', ``3.'', ``4.'' with ``\textasciitilde{}1\textasciitilde{}'', ``\textasciitilde{}2\textasciitilde{}'', ``\textasciitilde{}3\textasciitilde{}'', ``\textasciitilde{}4\textasciitilde{}''. We do this because the \textasciitilde{} character corresponds to the \texttt{<unknown>} token for T5 (Flan-T5 inherits the tokenizer directly from the original T5 model), clearly marking sentence boundaries as \texttt{<unknown><1><unknown>}, etc. In contrast, we observed that the original formatting might cause the tokenizer to merge subwords from before and after the markers with the bullet point itself, potentially resulting in ambiguous input for the generative models.

\paragraph{Fine-tuning} For fine-tuning PixArt-$\Sigma$ we use low-rank adapters (LoRAs)~\citep{hu2021loralowrankadaptationlarge} with a rank of 16 for 1 epoch on the fine-tuning dataset with a learning rate of $10^{-5}$, a cosine schedule with 1,000 warmup steps, and with an effective batch size of 640 text-image pairs. On the other hand, for fine-tuning Stable Diffusion Version~2 we employ a rank of 32 for 1 epoch, with a learning rate of $10^{-4}$, cosine schedule with 500 warmup steps, and an effective batch size of 320 samples.

\paragraph{Evaluating Text-Image Alignment} For text alignment evaluation we use two Visual Question Answering (VQA)~\citep{agrawal2016vqavisualquestionanswering} metrics: LLaVA-based \citep{liu2024improvedbaselinesvisualinstruction} and InstructBLIP-based \citep{dai2023instructblipgeneralpurposevisionlanguagemodels}. For each generated image–caption pair $(I_i, c_i)$ in our test set, we form a standardized yes/no query:
\[
\text{$Q_i$ = ``Is the figure showing: $c_i$?"}
\]
For each such query, the VQA models produces a probability for the answer ``yes". Thus, we report the average VQA score for the whole test set. We note that, given the longer length of our captions, we were unable to use standard CLIP-based text-image alignment metrics. Results are displayed in Table~\ref{tab:SD-finetuning} below.

\begin{table}[hbpt!]
\caption{Evaluation results comparing text-image alignment of fine-tuned models trained with structured versus shuffled image captions.}
  \resizebox{0.48\textwidth}{!}{%
    \begin{tabular}{|c|c|c|c|}
      \hline
      \textbf{Model} & \textbf{Captions}
        & \textbf{VQA LLaVA $\uparrow$} & \textbf{VQA InstructBLIP $\uparrow$}\\
      \hline
      \multirow{2}{*}{PixArt-$\Sigma$}
        & Structured   & \textbf{0.8630} & \textbf{0.8327} \\
        & Shuffled         & 0.8563 & 0.8303 \\
      \hline
      \multirow{2}{*}{Stable Diffusion 2}
        & Structured   & \textbf{0.8280} & \textbf{0.8140} \\
        & Shuffled         & 0.8120 & 0.8010 \\
      \hline
    \end{tabular}%
  }
  \label{tab:SD-finetuning}
\end{table}

We observe that fine-tuning on the full dataset requires many steps and can lead to image color saturation. According to guidelines for Stable Diffusion post-training, many practitioners recommend shorter fine-tuning durations to prevent overfitting and the aforementioned saturation issue. Hence, to minimize the effect of such confounding factors in our analysis we run an additional set of experiments in which we train for only 300 steps with ranks 4, 8, and 16. From Table~\ref{tab:SD-finetuning} and Table~\ref{tab:SD-finetuningextra} we can conclude that using structured captions consistently leads to better alignment.

\begin{table}[hbpt!]
\caption{Evaluation results comparing text-image alignment of fine-tuned models trained with structured versus shuffled image captions.}
  \resizebox{0.48\textwidth}{!}{%
    \begin{tabular}{|c|c|c|c|}
      \hline
      \textbf{Model} & \textbf{Captions}
        & \textbf{VQA LLaVA $\uparrow$} & \textbf{VQA InstructBLIP $\uparrow$}\\
      \hline
      \multirow{2}{*}{SD2 rank 4}
        & Structured   & \textbf{0.827} & \textbf{0.810} \\
        & Shuffled         & 0.825  & 0.809 \\
      \hline
            \multirow{2}{*}{SD2 rank 8}
        & Structured   & \textbf{0.828} & \textbf{0.811} \\
        & Shuffled         & 0.826 & 0.808 \\
      \hline
            \multirow{2}{*}{SD2 rank 16}
        & Structured   & \textbf{0.828} & \textbf{0.811} \\
        & Shuffled         & \textbf{0.828} & 0.810 \\
      \hline
    \end{tabular}%
  }
  \label{tab:SD-finetuningextra}
\end{table}

\section{Conclusion}
\label{sec:Conclusion}

The safety and reliability of generative models hinge significantly on the quality and structure of their training data. In this work, we argue that the challenge of achieving precise prompt adherence in text-to-image generative models often stems from the chaotic and unstructured nature of their training web-scale data. To address this issue, we propose the introduction of a canonical caption structure. This allows models to concentrate their learning capacity on semantic understanding rather than learning invariance to prompt reordering. We developed Re-LAION-Caption 19M, a refined dataset of 19 million high-resolution images. These images were re-captioned with LLaVA-Next to enforce a consistent, intuitive structure. Our preliminary experiments, fine-tuning models on both structured and randomly shuffled versions of our dataset, demonstrate compelling evidence that consistent captioning leads to significant improvements in prompt adherence. Looking ahead, this work opens several avenues for future research, including exploring different canonical structures, investigating the impact of varying degrees of structure on model performance, and applying this methodology to other modalities. By open-sourcing Re-LAION-Caption 19M, we hope to accelerate progress towards more reliable, predictable, and user-friendly generative AI.

We acknowledge that our re-captioning effort was conducted at a moderate scale—limited to 19 million images—due to computational constraints and a focus on fine-tuning. While scaling to billions of images may be impractical, we believe extending this approach to the scale of hundreds of millions remains feasible and could outperform training on larger datasets with unstructured captions. Furthermore, it would be valuable to explore the role of structured captions in other modalities, such as text-to-video generation.

\bibliography{example_paper}
\bibliographystyle{icml2025}

\clearpage
\appendix
\onecolumn
\section{Logbook of Procedure to Determine Accepted versus Rejected Samples}
\label{sec:Logbook}

In this section we walk the reader through the rationale for filtering the dataset. We aim to strike a balance between quality and retaining a high sample count.

\subsection{Aesthetic Score}

\begin{figure}[htbp!]
    \centering
\includegraphics[width=0.5\textwidth]{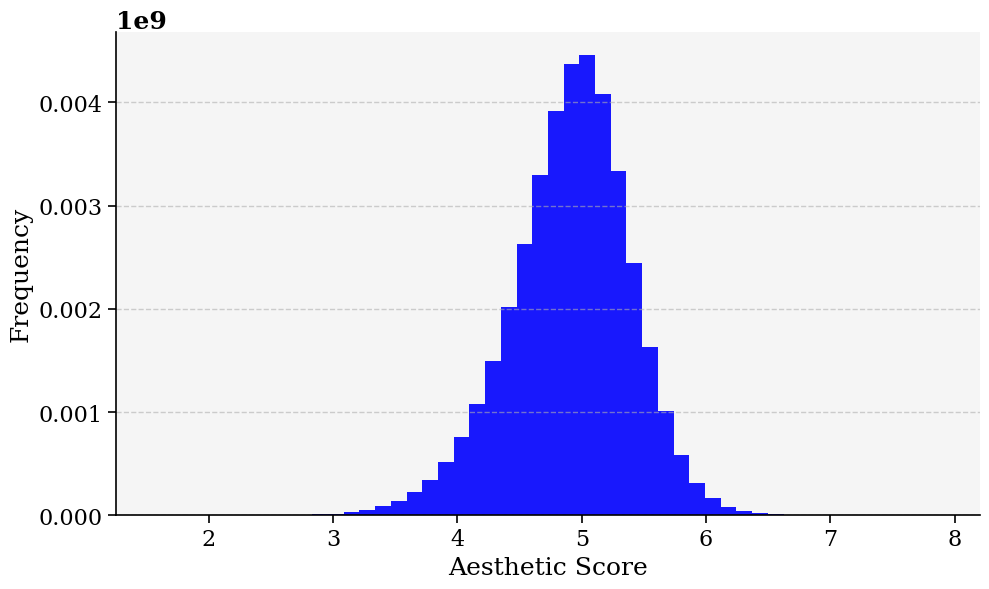}    \caption{Distribution of aesthetic score for the full 39 million subset.}
    \label{fig:aes_distput}
\end{figure}

\begin{figure}[htbp]
    \centering
    \includegraphics[width=\textwidth]{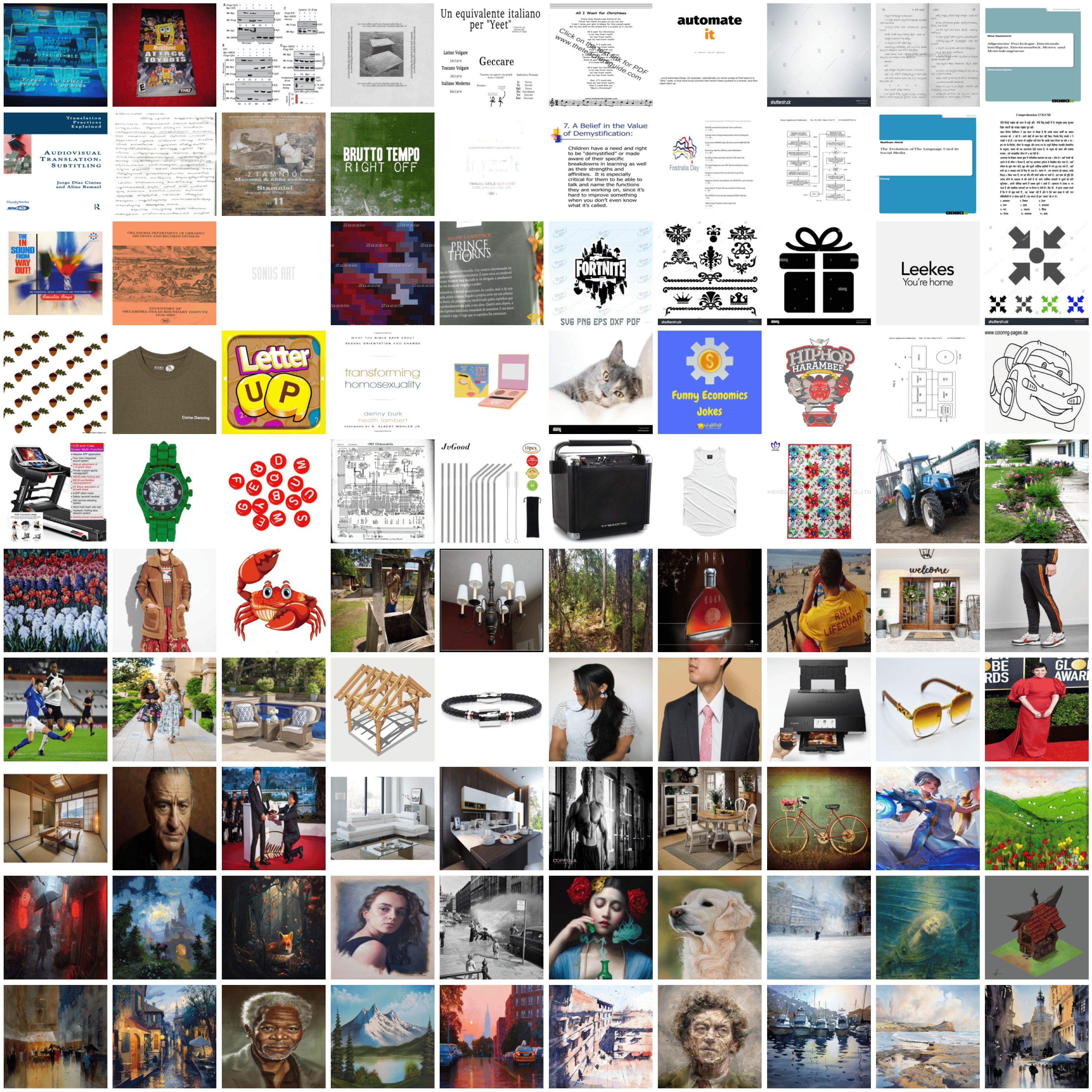}
    \caption{We subdivide the dataset into 10 buckets according to their mean aesthetic score. Each row corresponds to 10 samples from each of the buckets: lower to higher average aesthetic score from top to bottom.}
    \label{fig:aesthetic_score_samples}
\end{figure}

Figure~\ref{fig:aes_distput} displays the aesthetic score distribution for the 39,149,128 images remaining after the initial size and aspect ratio filtering previously mentioned in Section~\ref{sec:Filtering Scores}. This distribution resembles a Gaussian curve with the mean concentrated around a score of 5. Additionally, Figure~\ref{fig:aesthetic_score_samples} shows samples with varying aesthetic scores organized into 10 buckets: lower scores are assigned to samples at the top, while higher scores correspond to samples at the bottom. These samples clearly demonstrate that the aesthetic score generally correlates well with human perception. Lower scores are assigned to visually unappealing images with excessive text and lacking color, while higher scores are given to artistic or more professional-looking photographs and paintings. We also report the numerical values for mean and standard deviation scores per bucket and the sample count for each, see Table~\ref{tab:aesthetic_score_bins}. Note that the buckets are equal-width bins (or intervals). 

Samples in buckets 6 through 10 exhibit demonstrably high quality. Specifically, they feature a clear focus on the main subject, appealing color distributions, an absence of random text or scribbles, and a noticeable artistic touch. On the contrary, buckets 1 through 5 present a high count of suboptimal samples, so we discard them. Based on this, we retain 26,457,998 samples from the original 39,149,128. This is the most aggressive filtering step, after the preliminary size based filtering.

\begin{table}[htbp]
  \centering
  \caption{Aesthetic Score Distribution Segmented into 10 buckets}
  \scalebox{0.8}{
  \begin{tabular}{@{}lccc@{}}
    \toprule
    \textbf{Bucket} & \textbf{Mean} & \textbf{Standard Deviation} & \textbf{Count} \\
    \midrule
    1 &  2.05 & 0.13 & 261 \\
    2 & 2.67 & 0.14 &  9,328\\
    3 &  3.28 & 0.15 & 189,482 \\
    4 &  3.88 & 0.16 & 1,982,559 \\
    5 & 4.48 & 0.17 &  10,509,500\\
    6 &  5.03 & 0.17 & 20,153,707 \\
    7 & 5.56  & 0.15 & 5,980,468 \\
    8 & 6.15 & 0.14 & 314,390 \\
    9 & 6.78  & 0.13 &  9,293\\
    10 & 7.42  & 0.15 & 140 \\
    \bottomrule
  \end{tabular}}
\label{tab:aesthetic_score_bins}
\end{table}

\clearpage
\subsection{Luminance}

\begin{figure}[htbp!]
    \centering
\includegraphics[width=0.5\textwidth]{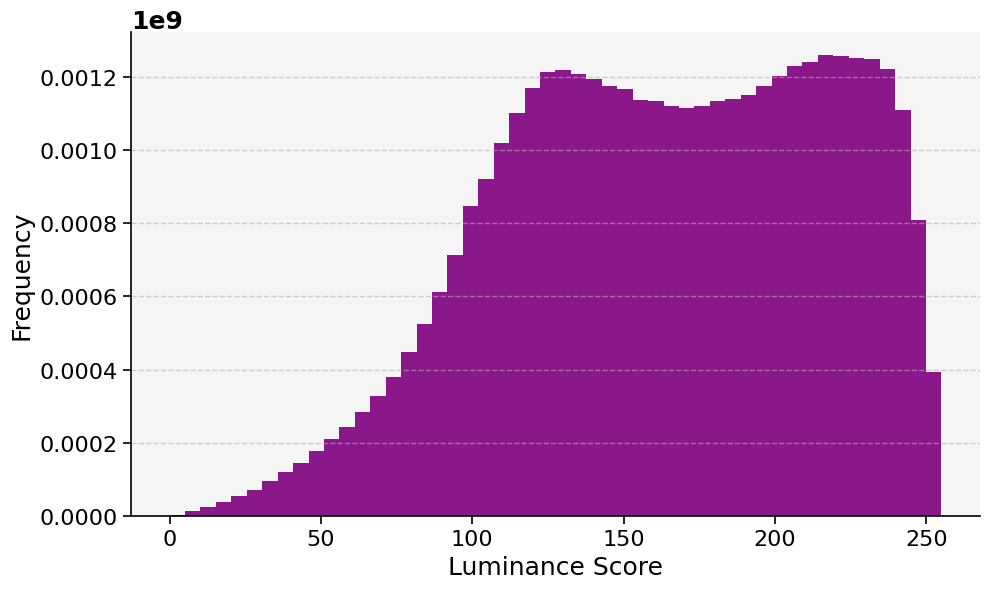}    \caption{Distribution of luminance score for the full 39 million subset.}
    \label{fig:luminance_dist}
\end{figure}

In Figure~\ref{fig:luminance_dist}, we plot the luminance score distribution for the dataset. As before, we initially subdivided this distribution into 10 buckets. However, we saw that many samples exhibited a relatively high luminance score, which we hypothesize may be attributed to the presence of a large number of images from online stores where products are displayed against a white background. We found these samples to be less interesting.

\begin{figure}[htbp]
    \centering
    \includegraphics[width=0.65\textwidth]{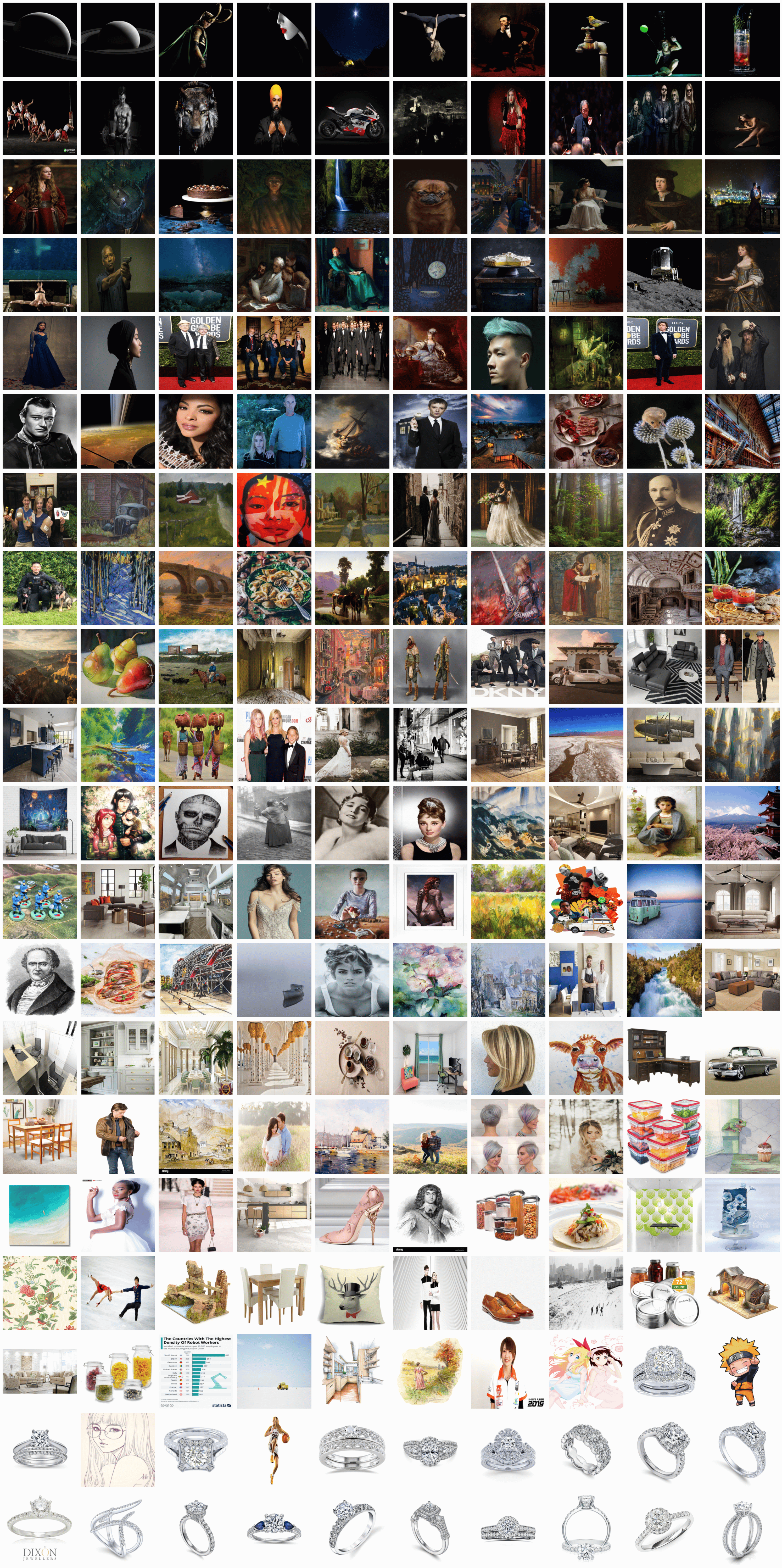}
    \caption{We subdivide the dataset into 20 buckets according to their mean luminance score for images with an aesthetic score higher than 6 only. Each row corresponds to 10 samples from each of the buckets: lower to higher average aesthetic score from top to bottom.}
    \label{fig:luminance_examples_ae6}
\end{figure}

\begin{figure}[htbp]
    \centering
    \includegraphics[width=0.65\textwidth]{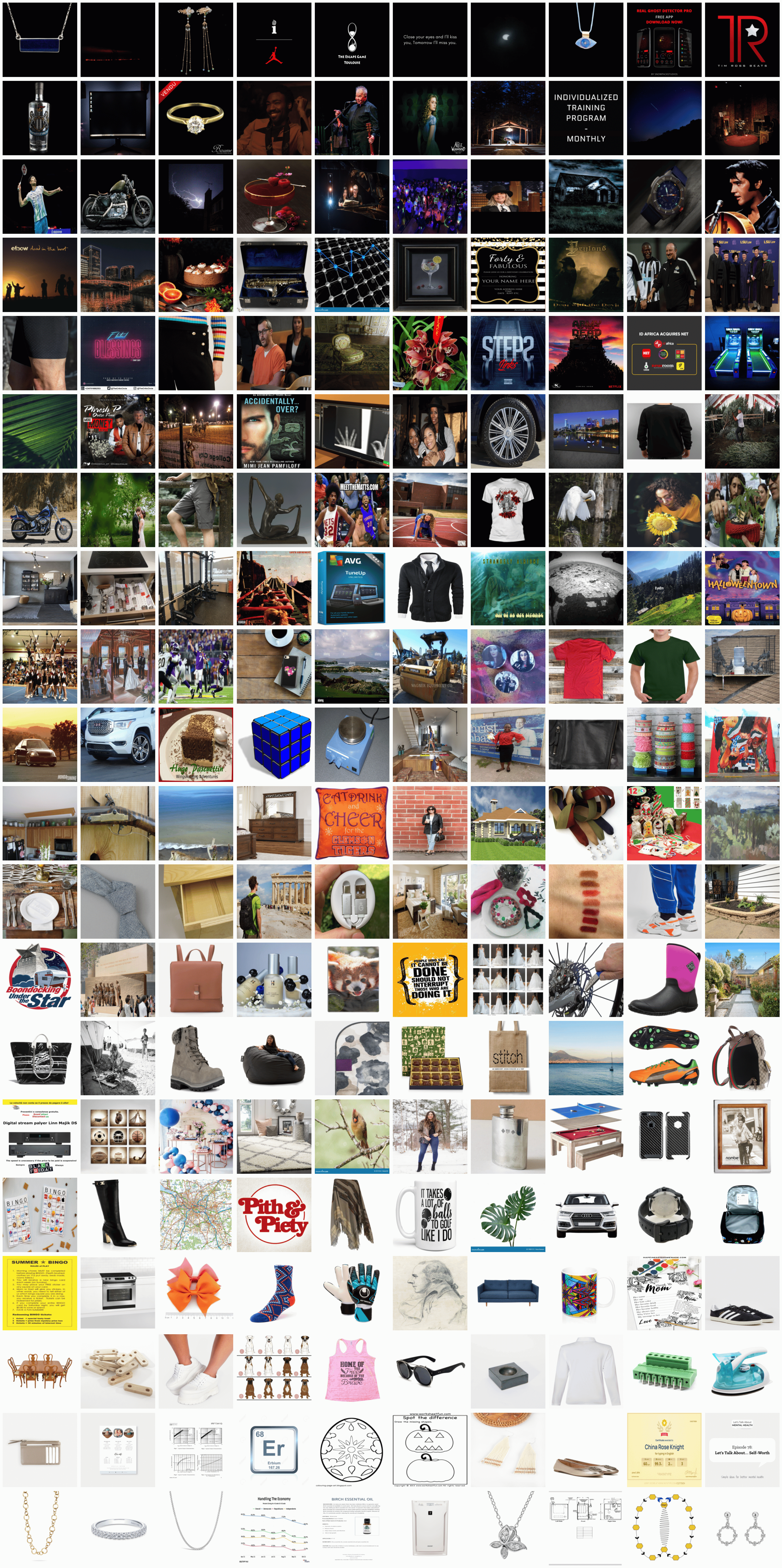}
    \caption{We subdivide the dataset into 20 buckets according to their mean luminance score for the full 39 million image subset.}
    \label{fig:luminance_examples_any_ae}
\end{figure}

To verify our hypothesis, we subdivide images with aesthetic scores above 6 into 20 luminance buckets (Figure~\ref{fig:luminance_examples_ae6}). Even in these high-scoring examples, the highest luminance buckets contain product images. Interestingly we note a high presence of diamond rings. This suggests the aesthetic score can be inflated by the presence of an isolated, visually appealing object, potentially overlooking the overall image context. For instance, it might score a close-up of a `pretty' ring on a plain background highly, similar to a well-composed portrait. The general luminance distribution (Figure~\ref{fig:luminance_examples_any_ae}) also shows a strong correlation between high luminance and product pictures. 

Based on this observation we filter out the four highest luminance buckets, see Table~\ref{tab:luminance_score_bins}, to avoid including an excessive count of online products. Likewise we also filter the darkest bucket containing approximately 30,000 images to prevent our model from producing an excessive amount of black images. However, in this case we filter less aggressively than for high luminance since we observe a lot of eye-pleasing dark samples. This reduces the effective dataset from 39,149,128 to 28,108,375, and with the joint filtering with aesthetic score the number goes down to 19,858,589 images.

\begin{table}[htbp]
  \centering
  \caption{Luminance Distribution Segmented into 20 Buckets}
  \scalebox{0.7}{
  \begin{tabular}{@{}lccc@{}}
    \toprule
    \textbf{Bucket} & \textbf{Mean} & \textbf{Standard Deviation} & \textbf{Count} \\
    \midrule
    1 & 8.64 & 2.94 & 31,203 \\
    2 & 20.05 & 3.58 & 109,466 \\
    3 & 32.57 & 3.62 & 226,347 \\
    4 & 45.12 & 3.66 & 388,026 \\
    5 & 57.76 & 3.68 & 591,275 \\
    6 & 70.50 & 3.68 & 857,594 \\
    7 & 83.29 & 3.66 & 1,266,129 \\
    8 & 96.03 & 3.65 & 1,875,938 \\
    9 & 108.62 & 3.66 & 2,477,027 \\
    10 & 121.24 & 3.67 & 2,939,717 \\
    11 & 133.85 & 3.68 & 3,023,074 \\
    12 & 146.59 & 3.68 & 2,934,816 \\
    13 & 159.36 & 3.68 & 2,830,559 \\
    14 & 172.13 & 3.68 & 2,791,116 \\
    15 & 184.89 & 3.68 & 2,846,527 \\
    16 & 197.69 & 3.68 & 2,950,764 \\
    17 & 210.41 & 3.69 & 3,098,098 \\
    18 & 223.12 & 3.68 & 3,134,427 \\
    19 & 235.77 & 3.66 & 3,045,293 \\
    20 & 247.12 & 3.30 & 1,731,732 \\
    \bottomrule
  \end{tabular}}
\label{tab:luminance_score_bins}
\end{table}

\clearpage
\subsection{OCR Score}

\begin{figure}[htbp!]
    \centering
\includegraphics[width=0.5\textwidth]{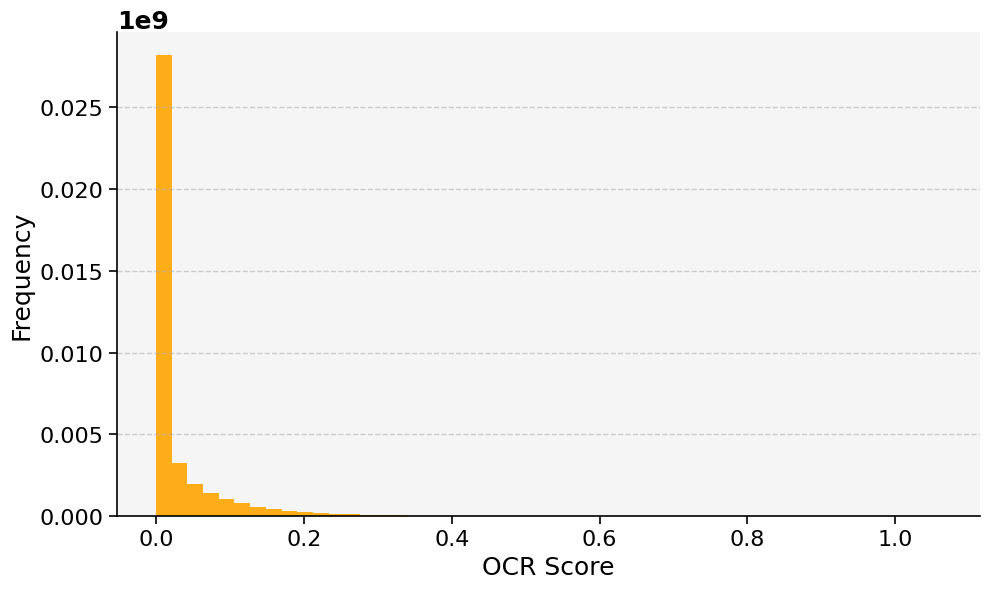}    \caption{Distribution of scores.}
    \label{fig:ocr_dist}
\end{figure}

\begin{figure}[htbp]
    \centering
    \includegraphics[width=\textwidth]{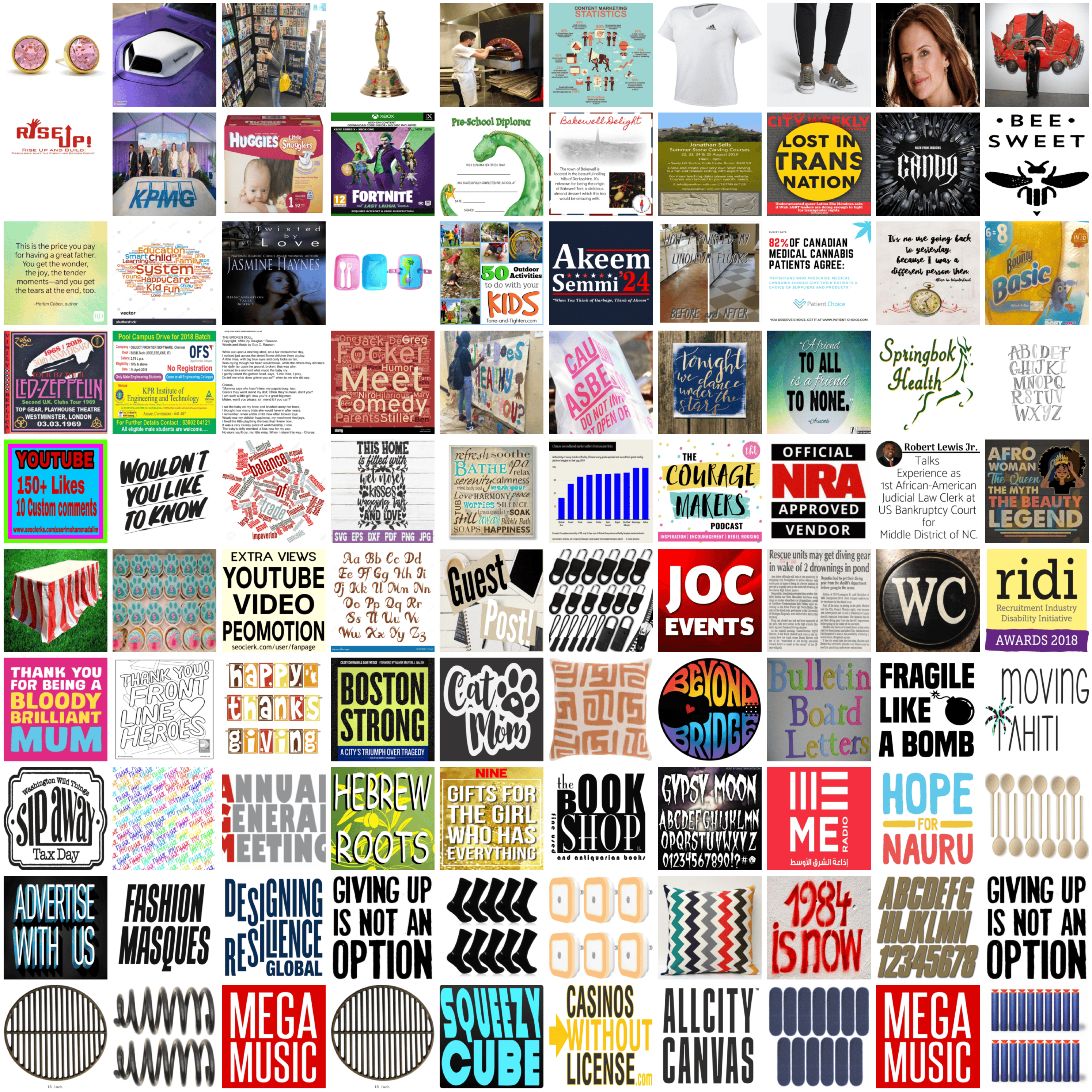}
    \caption{We subdivide the dataset into 10 buckets according to their OCR score. Each row corresponds to 10 samples from each of the buckets: lower to higher average OCR score from top to bottom.}
    \label{fig:ocr_viz}
\end{figure}

For OCR filtering, we retain buckets 1, 7, 8, 9, and 10 (see Table~\ref{tab:ocr_score_bins}). Bucket 1 contains most images due to the skewed distribution (Figure~\ref{fig:ocr_dist}). Intermediate buckets primarily include scribbles or text too small for LLaVA models (Figure~\ref{fig:ocr_viz}), while only the last few buckets feature sufficiently large and clear text beneficial for downstream generative model training. The OCR filtering by itself reduces the dataset from 39,149,128 to 35,882,851, and applying all filters in conjunction we obtain the final Re-LAION-Caption 19M subset with a total of 19,055,277 images.

\begin{table}[htbp]
  \centering
  \caption{OCR score Distribution Segmented into 10 buckets}
  \scalebox{0.8}{
  \begin{tabular}{@{}lccc@{}}
    \toprule
    \textbf{Bucket} & \textbf{Mean} & \textbf{Standard Deviation} & \textbf{Count} \\
    \midrule
    1 & 0.01 & 0.02 & 35,878,834 \\
    2 & 0.15 & 0.03 & 2,428,923 \\
    3 & 0.25 & 0.03 & 629,186 \\
    4 & 0.36 & 0.03 & 155,339 \\
    5 & 0.47 & 0.03 & 41,369 \\
    6 & 0.57 & 0.03 & 11,460 \\
    7 & 0.68 & 0.03 & 3,047 \\
    8 & 0.78 & 0.03 & 789 \\
    9 & 0.89 & 0.03 & 157 \\
    10 & 1.00 & 0.03 & 24 \\
    \bottomrule
  \end{tabular}}
\label{tab:ocr_score_bins}
\end{table}

\clearpage

\section{Visualizing the Final Re-LAION-Caption 19M Subset}
\label{sec:Visualizing}

\begin{figure}[htbp!]
    \centering
\includegraphics[width=0.5\textwidth]{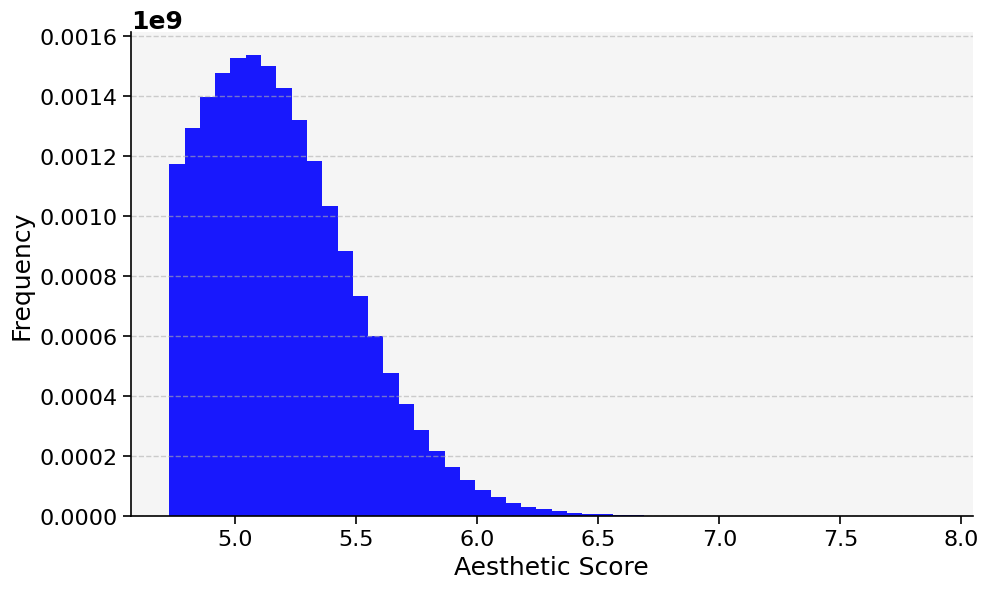}    \caption{Distribution of aesthetic scores for Re-LAION-Caption 19M Subset.}
    \label{fig:ae_dist_Re-LAION-Caption}
\end{figure}

\begin{figure}[htbp!]
    \centering
\includegraphics[width=0.5\textwidth]{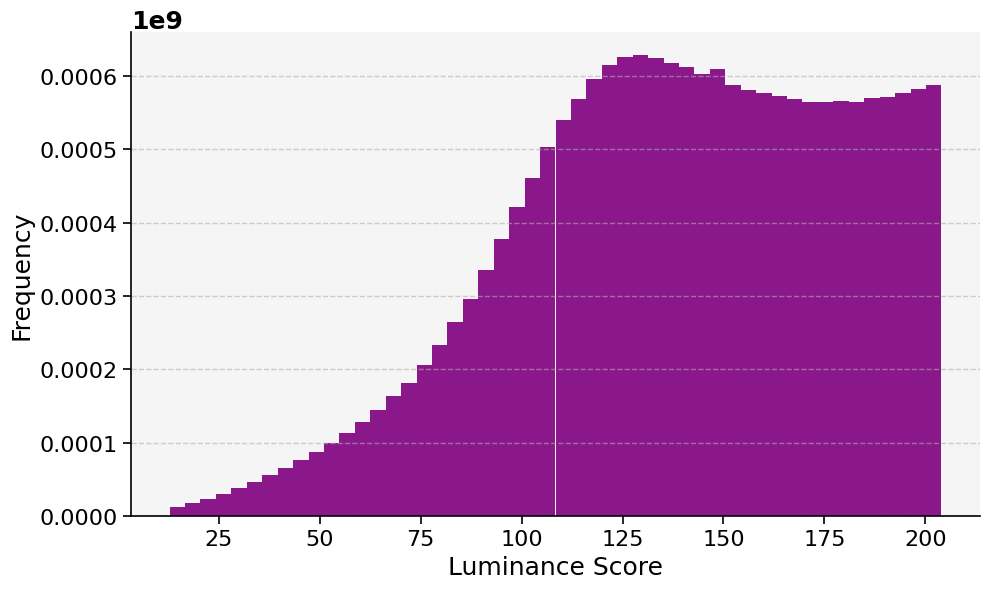}    \caption{Distribution of luminance scores for Re-LAION-Caption 19M Subset.}
    \label{fig:lum_dist_Re-LAION-Caption}
\end{figure}

\begin{figure}[htbp!]
    \centering
\includegraphics[width=0.5\textwidth]{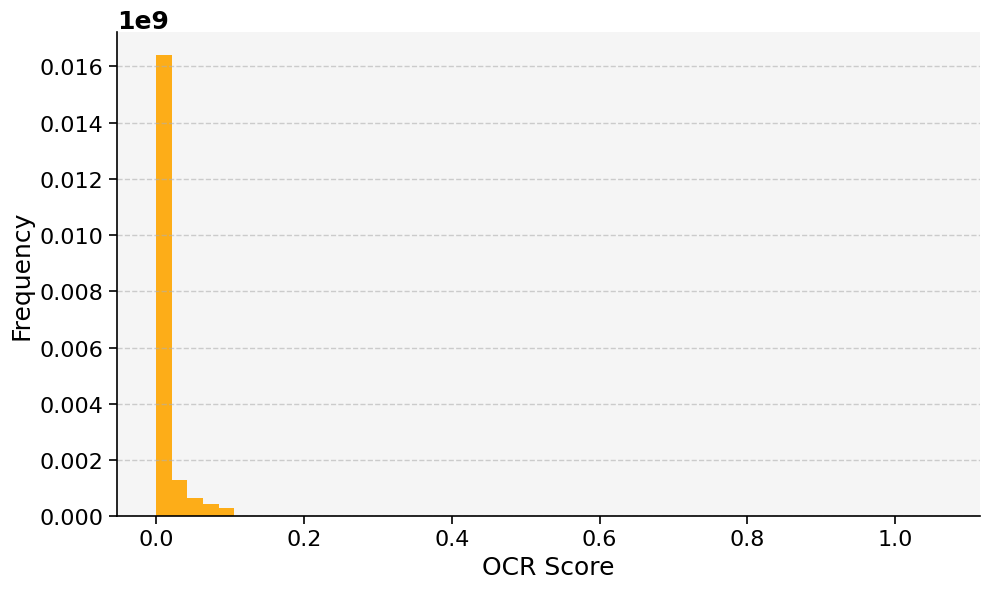}    \caption{Distribution of OCR scores for Re-LAION-Caption 19M Subset.}
    \label{fig:ocr_dist_Re-LAION-Caption}
\end{figure}

\begin{figure}[htbp!]
    \centering
    \includegraphics[width=\textwidth]{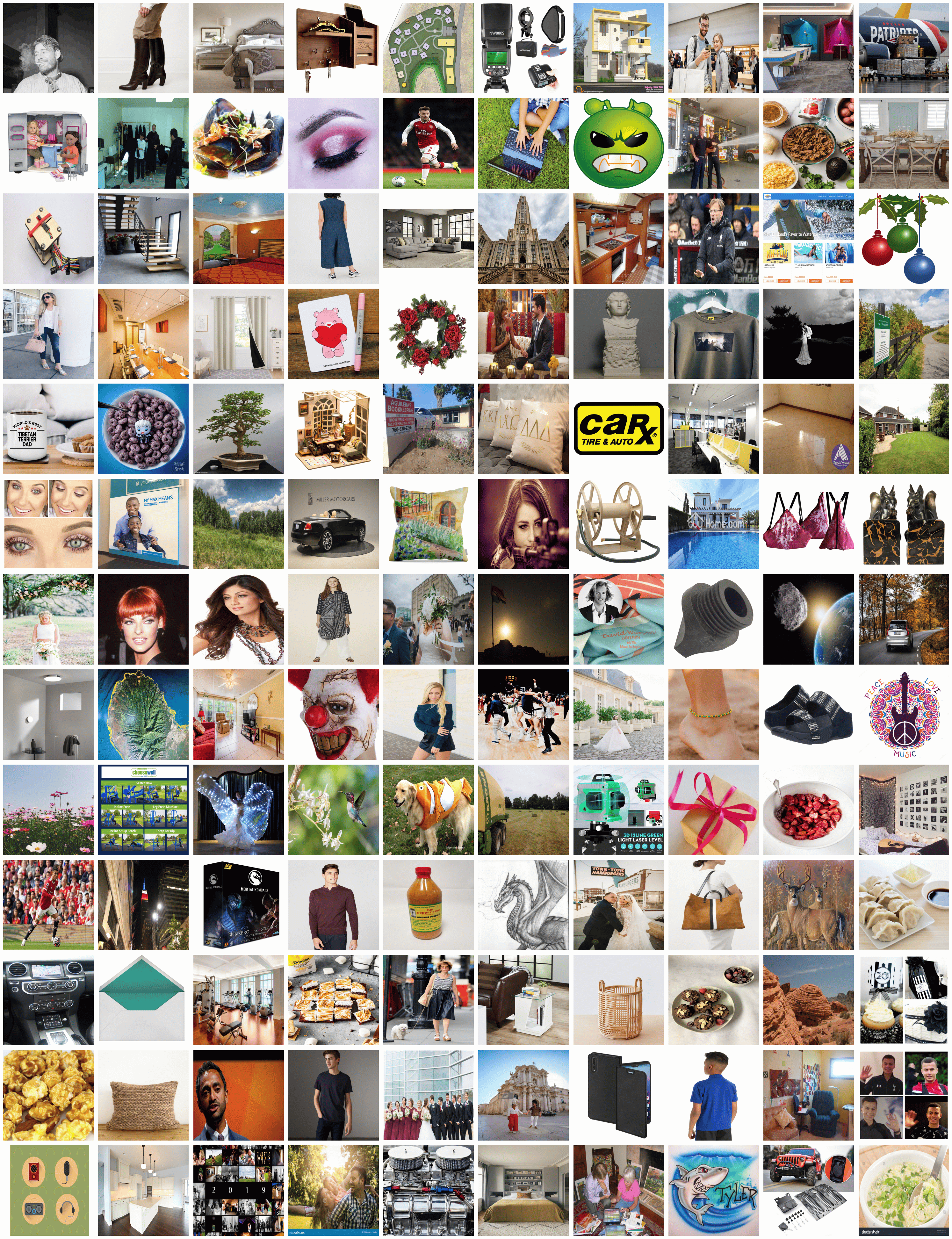}
    \caption{Random set of samples from Re-LAION-Caption 19M}
    \label{fig:caption_subset1}
\end{figure}

\begin{figure}[htbp!]
    \centering
    \includegraphics[width=\textwidth]{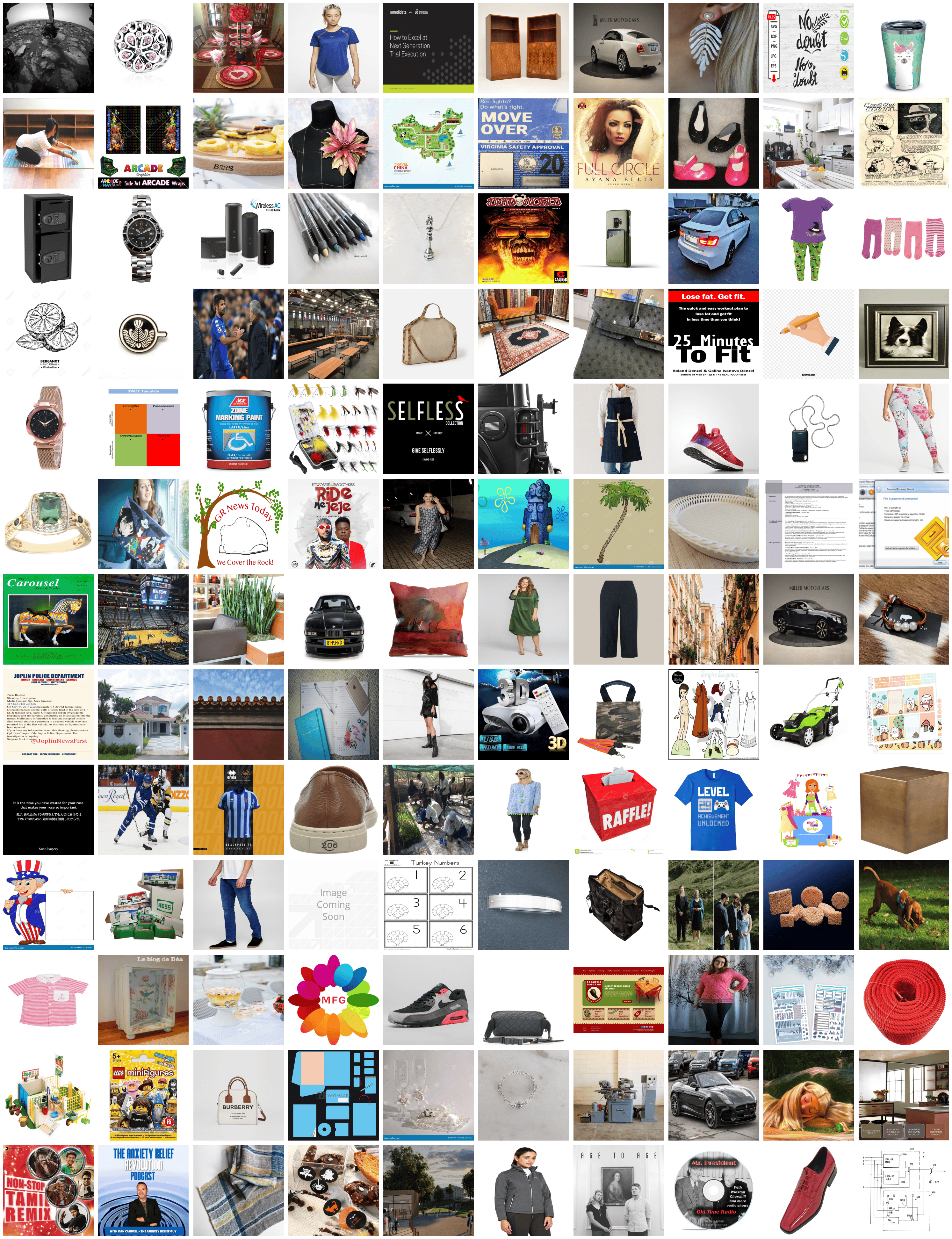}
    \caption{Random set of samples from original Re-LAION dataset}
    \label{fig:original_subset1}
\end{figure}

\begin{figure}[htbp!]
    \centering
\includegraphics[width=0.65\textwidth]{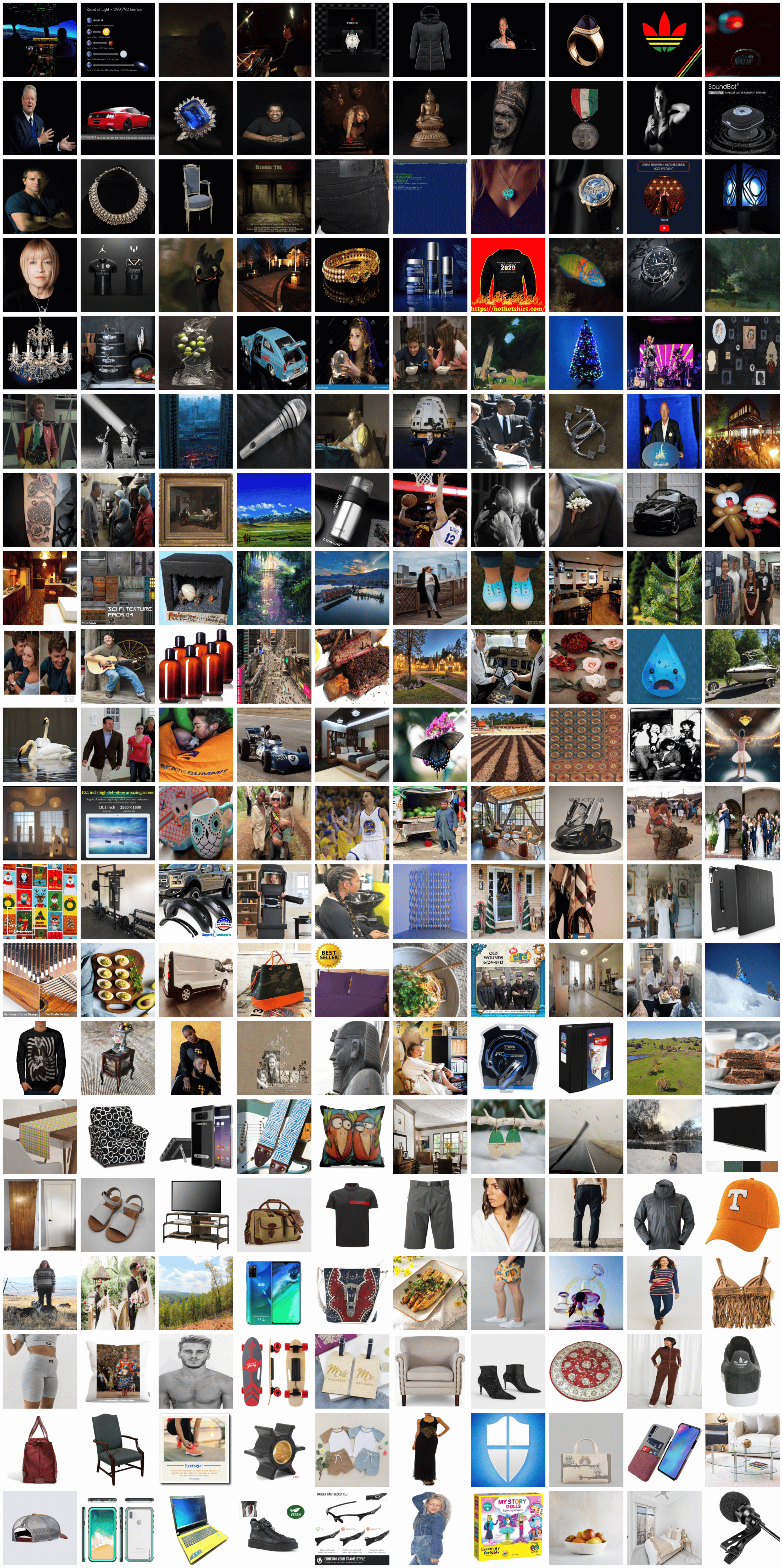}    \caption{Luminance buckets for Re-LAION-Caption 19M Subset.}
    \label{fig:filt_lum_20}
\end{figure}

We visualize the distributions for our new dataset after applying all filters together in Figure~\ref{fig:ae_dist_Re-LAION-Caption},~\ref{fig:lum_dist_Re-LAION-Caption}, and~\ref{fig:ocr_dist_Re-LAION-Caption}. Additionally we compare random samples from Re-LAION-Caption 19M in Figure~\ref{fig:caption_subset1}, against those in the original Re-LAION dataset in Figure~\ref{fig:original_subset1}. We can clearly see that our dataset minimizes the presence of online products, has a higher density of aesthetically pleasing images, and generally only contains legible text while minimizing the presence of scribbles and small fonts. Lastly, in Figure~\ref{fig:filt_lum_20} we show that for Re-LAION-Caption 19M the high luminance buckets have a lower likelihood to display images corresponding to online products that we have previously observed in Figure~\ref{fig:luminance_examples_any_ae}.

\clearpage

% \begin{figure}[htbp!]
%     \centering
%     \includegraphics[width=\textwidth]{figures/filtered_viz_3.png}
%     \caption{Additional random set of samples from Re-LAION-Caption 30M}
%     \label{fig:caption_subset2}
% \end{figure}

\clearpage

\section{LLaVa captioning failure cases}
\label{sec:LLaVa captioning failure cases}

\begin{figure*}[htbp!]
    \centering
    \includegraphics[width=\textwidth,trim= 0 0 0 0,clip]{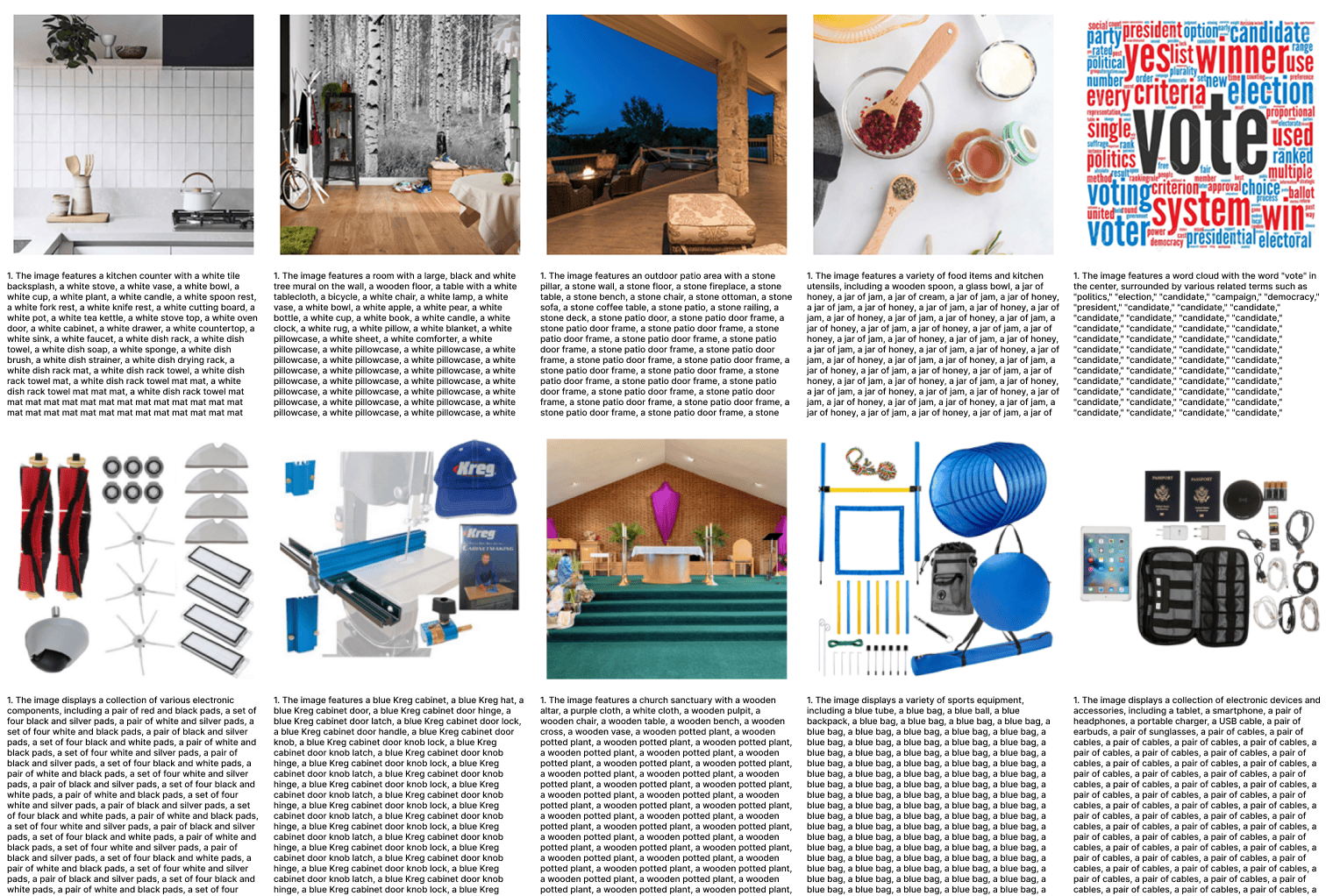}
    \caption{Defective text-image pairs identified when re-captioning images using Mistal 7B Instruct-based LlaVa-Next. Note that captions are cropped due to their excessive length.}
    \label{fig:llava_failure_examples}
\end{figure*}

\begin{figure*}[htbp!]
    \centering
    \includegraphics[width=\textwidth,trim= 0 30 0 0,clip]{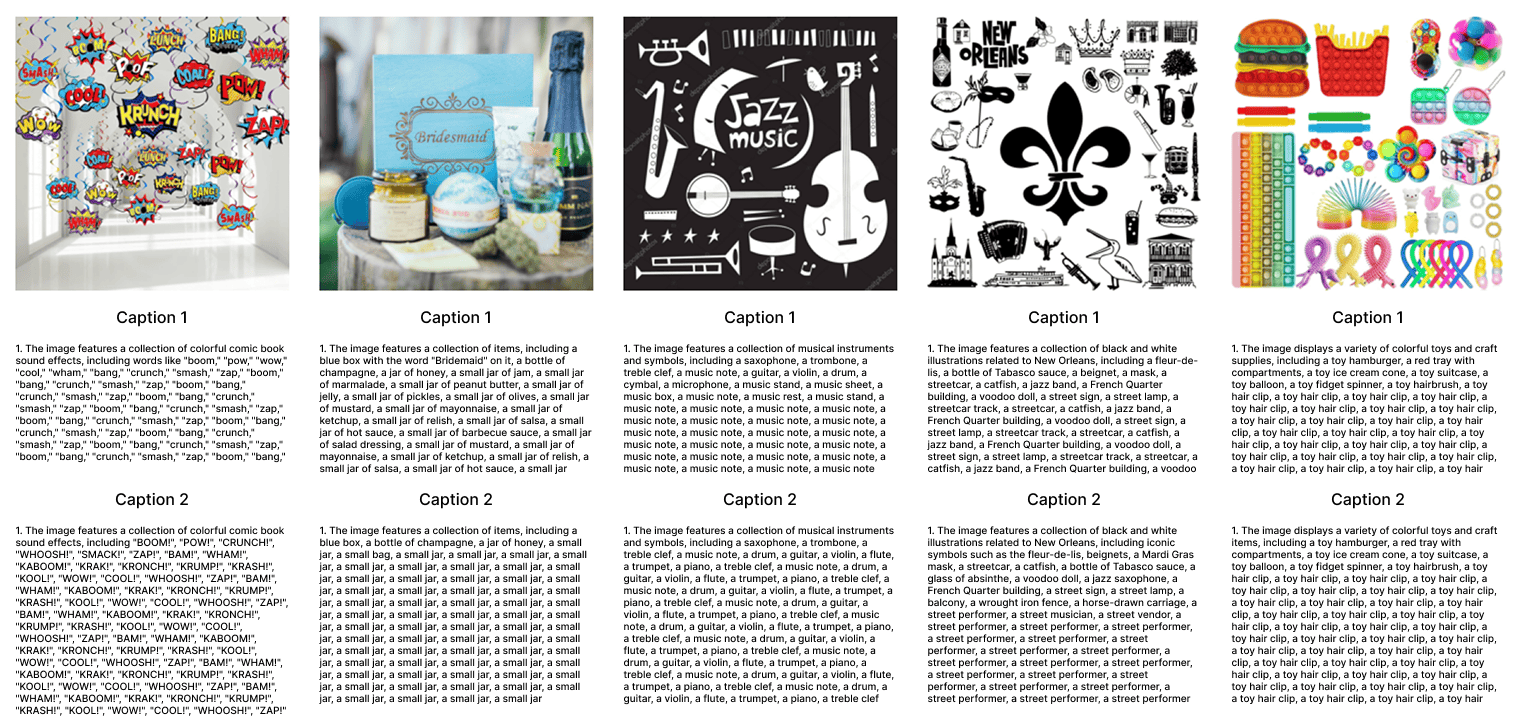}
    \caption{Examples of persistent defective captioning by LLaVa-Next.}
    \label{fig:multiple_defective}
\end{figure*}

We present examples of defective captions produced by LLaVa-Next in Figure~\ref{fig:llava_failure_examples}: the model seems to get stuck in an infinite loop when trying to describe images with many elements. After rerunning the captioning process on the 17,320 defective samples initially identified, we found that this resolved only 122 text-image pairs. As shown in Figure~\ref{fig:multiple_defective}, some images appear particularly challenging for LLaVa-Next, and even with different seeds, the model consistently struggles to generate good captions. Therefore, we discarded the remaining 17,198 defective samples, resulting in a final dataset comprising 19,038,079 text-image pairs.

\end{document}